\PassOptionsToPackage{sort&compress}{natbib}
\documentclass[final,3p,12pt]{elsarticle}
\usepackage{multirow}
\makeatletter
\def\ps@pprintTitle{%
  \let\@oddhead\@empty
  \let\@evenhead\@empty
  \def\@oddfoot{\reset@font\hfil\thepage\hfil}
  \let\@evenfoot\@oddfoot
}
\makeatother

\usepackage{amssymb}
\usepackage{tabularray}
\usepackage{placeins}
\usepackage{adjustbox} 
\usepackage{booktabs}
\usepackage{amsmath}
\usepackage{lineno}
\usepackage{makecell} 
\usepackage{xcolor}
\usepackage{graphicx}
\usepackage{rotating}
\usepackage{afterpage}
\usepackage{subfigure}
\usepackage{pdflscape}
\usepackage{xcolor}
\usepackage{colortbl}
\usepackage{setspace}
\usepackage{placeins}
\usepackage{array} 
\begin{document}

\begin{frontmatter}


\title{Discovery of Spatter Constitutive Models in Additive Manufacturing Using Machine Learning}

\author[inst1]{Olabode T. Ajenifujah}
\affiliation[inst1]{organization={Department of Mechanical Engineering, Carnegie Mellon University},
            city={Pittsburgh},
            postcode={15213}, 
            state={PA},
            country={USA}}
            



\author[inst1,inst3,inst4]{Amir Barati Farimani\corref{corauthor}}
\affiliation[inst3]{organization={Department of Chemical Engineering, Carnegie Mellon University},
            city={Pittsburgh},
            postcode={15213}, 
            state={PA},
            country={USA}}
\affiliation[inst4]{organization={Machine Learning Department, Carnegie Mellon University},
            city={Pittsburgh},
            postcode={15213}, 
            state={PA},
            country={USA}}

\begin{abstract}
Additive manufacturing (AM) is a rapidly evolving technology that has attracted applications across a wide range of fields due to its ability to fabricate complex geometries. However, one of the key challenges in AM is achieving consistent print quality. This inconsistency is often attributed to uncontrolled melt pool dynamics, partly caused by spatter which can lead to defects. Therefore, capturing and controlling the evolution of the melt pool is crucial for enhancing process stability and part quality. In this study, we developed a framework to support decision-making towards efficient AM process operations, capable of facilitating quality control and minimizing defects via  machine learning (ML) and polynomial symbolic regression models. We implemented experimentally validated computational tools, specifically for  laser powder bed fusion (LPBF) processes as a cost-effective approach to collect large datasets. For a dataset consisting of 281 varying process conditions, parameters such as melt pool dimensions (length, width, depth), melt pool geometry (area, volume), and volume indicated as spatter were extracted. Using machine learning (ML) and polynomial symbolic regression models, a high R\(^2\) of over 95 \% was achieved in predicting the melt pool dimensions and geometry features on both the training and testing datasets, with either process conditions (power and velocity) or melt pool dimensions as the model inputs. In the case of volume indicated as spatter, the value of the R\(^2\) improved after logarithmic transforming the model inputs, which were either the process conditions or the melt pool dimensions. Among the investigated ML models, the ExtraTree model achieved the highest R\(^2\) values of 96.7 \% and 87.5 \%. With respect to the symbolic regression model, R\(^2\) values of 85 \% and 82 \% were achieved on the training and testing datasets, respectively. Our study culminated in the discovery of symbolic equations based on model inputs in the polynomial fitting model for all investigated parameters, thereby providing interpretable evaluations of the feature importance.



\end{abstract}

\begin{keyword}
 Additive Manufacturing \sep Symbolic Regression\sep  Process Map\sep Machine Learning \sep Defects. 
\end{keyword}

\end{frontmatter}


\section{Introduction}

Laser Powder Bed Fusion (LPBF) is a popular metal additive manufacturing (AM) process that has demonstrated high precision and remarkable performance in producing complex part geometries and internal structures with micron-scale precision. The operating process of Laser Powder Bed Fusion (LPBF) involves guiding a laser using a scanning galvanometer to selectively melt a layer of powder deposited on a build plate. Once solidification occurs, the next layer of powder is spread and the process is repeated iteratively until the final part is complete \cite{li2022convolutional}. In addition, LPBF offers significant advantages, including design flexibility, an automated method for building parts directly at their point of use from CAD files, and reduced material waste. These benefits have facilitated its adoption in various industries, including aerospace, automotive, medicine, energy, and prototyping \cite{wirth2022implementation, repossini2017use}. However, despite its increasing application in different sectors, compared to other forms of manufacturing processes such as rolling, casting, and machining, LPBF still occupies a low market share, which is attributed to its low rate of part production and poor quality control. A popular approach to remediate low part production involves using multi-lasers, which, however, increases the defect formation rate. The efficacy of state-of-the-art methods for quality control is assessed by comparing the properties and performance of the finished part with the target standard, although variability and suboptimal performance are still observed.

Identifying optimal process conditions typically relies on costly and time-intensive trial-and-error experiments across a wide range of process parameters. This challenge has motivated the use of 3D numerical modeling tools, which can effectively capture the complex phenomena occurring during melting and provide insights into parameters that are difficult or impossible to measure using current in situ experimental techniques. However, a high-fidelity AM simulation package that will account for all of the complex physics in AM is computationally expensive. As a result, increasing attention is being directed towards a robust smart AM manufacturing tool, which has been dubbed a solution to match the desired properties and performance of parts and AM tools \cite{xiao2024data}. This approach involves developing a digital twin of the LPBF process, which will be capable of mirroring the real-time evolution of part-building processes, displaying key measurements, and forecasting future events during laser-material interactions. This type of system will enhance the detection of faults or anomalies during the printing process through the integration of a sensitive control system to mitigate or prevent the intricacies that may occur during the printing process. The core component of the digital twin will be the integration of experimental datasets, system modeling analysis, and ML models. ML enables predictive and performance insights that can be difficult to uncover through traditional physics-based modeling approaches due to the high computation cost. \cite{ajenifujah2024integrating, knapp2017building, zhang2020digital, wang2020investigation, ogoke2024inexpensive}. Monitoring the formation and dynamic processes that occur in the melt pool is one of the crucial aspects of LBPF operation that can improve quality control by providing an understanding of the underlying processes that take place in relation not only to the process conditions, but also to the material types that govern the nature of the melt pool, the microstructure, and the mechanical properties \cite{wirth2022implementation}. Consequently, defect formation, such as porosity, surface roughness, residual stresses, warping, cracking, and delamination that can influence quality control, will be further elucidated and prevented. 

The general characteristics of the melt pool that have been popularly monitored to understand the formation of defects or the properties of built parts are dimensions \cite{cunningham2017synchrotron, chen2020melt, kiss2019laser, guo2019situ, tang2024role}, shape\cite{leach2019geometrical,reijonen2024effect, liao2024deep, farimani2024llm, scime2019melt}, temperature distribution \cite{yan2018review, myers2023high, myers2023two, barua2014vision, kuriya2018relationship, hemmasian2023surrogate, khanzadeh2018dual}, oxygen content \cite{chia2024unveiling, eo2021melt, calta2020pressure, leung2019effect, soundarapandiyan2023situ}, and pressure \cite{calta2020pressure}. Many studies have been reported on coupling ML with monitoring parameters collected either through experimental or modeling methods to accelerate the interpretation of system behavior. Xiao et al. \cite{xiao2024data} predicted the future melt pool area using the history of the previous construction. They first filtered out noise using a convolutional neural network (CNN)-based model, then quantitatively estimated future melt pool areas using an artificial neural network (ANN) trained on scanning history, particularly past melt pools. This approach significantly reduced the average relative error magnitude (AREM) to 2.8\% compared to 14.8\% with the existing Neighboring Effect Modeling Method (NBEM). Akbari et al. \cite{akbari2022meltpoolnet} predicted the dimensions of the melting pool using ML models, where the process parameters and the material properties are curated from different source of published literatures. Zhang et al.  \cite{zhang2024machine} implemented LSTM-based approach to estimate the area of the meltpool with accuracy of 90.7 \%, then used Melt Pool Generative Adversarial Network to synthesize the images of the melt pool and achieve a structural similarity score of 0.91. Wang et al. \cite{wang2024sub} implemented a machine learning-assisted approach based on a deep neural network to demodulate the optical signal to thermal distribution and significantly improve spatial resolution to 28.8 µm/pixel spatial resolution and 10 kHz sampling frequency, ideal for measuring the sharp thermal gradient and cooling rates in the L-PBF process.  Scime et al. \cite{scime2019using}  used a high-speed camera to capture the images of the molten pool, then proposed an unsupervised learning algorithm to distinguish the molten pool and identify defects.

Extending quality control to the monitoring of the ejection of the spatter from the melt pool can provide information on the state or stability of the melt pool as defects develop \cite{andani2017spatter, andani2018study, criales2017laser, esmaeilizadeh2019effect}. In-homogeneity fusion and thermal gradient caused by spatter can be a source of crack initiation sites \cite{otegui1989fatigue}. The traveling spatter can absorb the laser radiation or reflect it, causing a fluctuation in the intensity of the laser \cite{ye2019investigation}. Limited research has explored the integration of machine learning (ML) models with spatter formation. Most studies in this area have focused mainly on spatter detection by coupling ML with computer vision or by classifying spatter according to process conditions or signals \cite{repossini2017use, chebil2023deep, cai2024real, wu2022online, wang2021characteristics, o2024computational}. However, these approaches do not explain the underlying mechanisms that govern the ejection rate of the spatter, which fluctuates according to the process conditions and the state of the melt pool. Previously, we implemented machine learning (ML) to classify and detect slight variations in the measured parameters between the spatter and the melt pool at 5 µs. This approach revealed the ranking of different measured parameters based on their importance in spatter ejection. To enhance interpretability, we applied an explainable AI technique known as SHapley Additive exPlanations (SHAP), which allowed us to identify the range of the feature values that positively contribute to spatter predictions \cite{ajenifujah2024integrating}.

\begin{figure*}[htbp!]
\begin{center}
\includegraphics[width=1.1\linewidth]{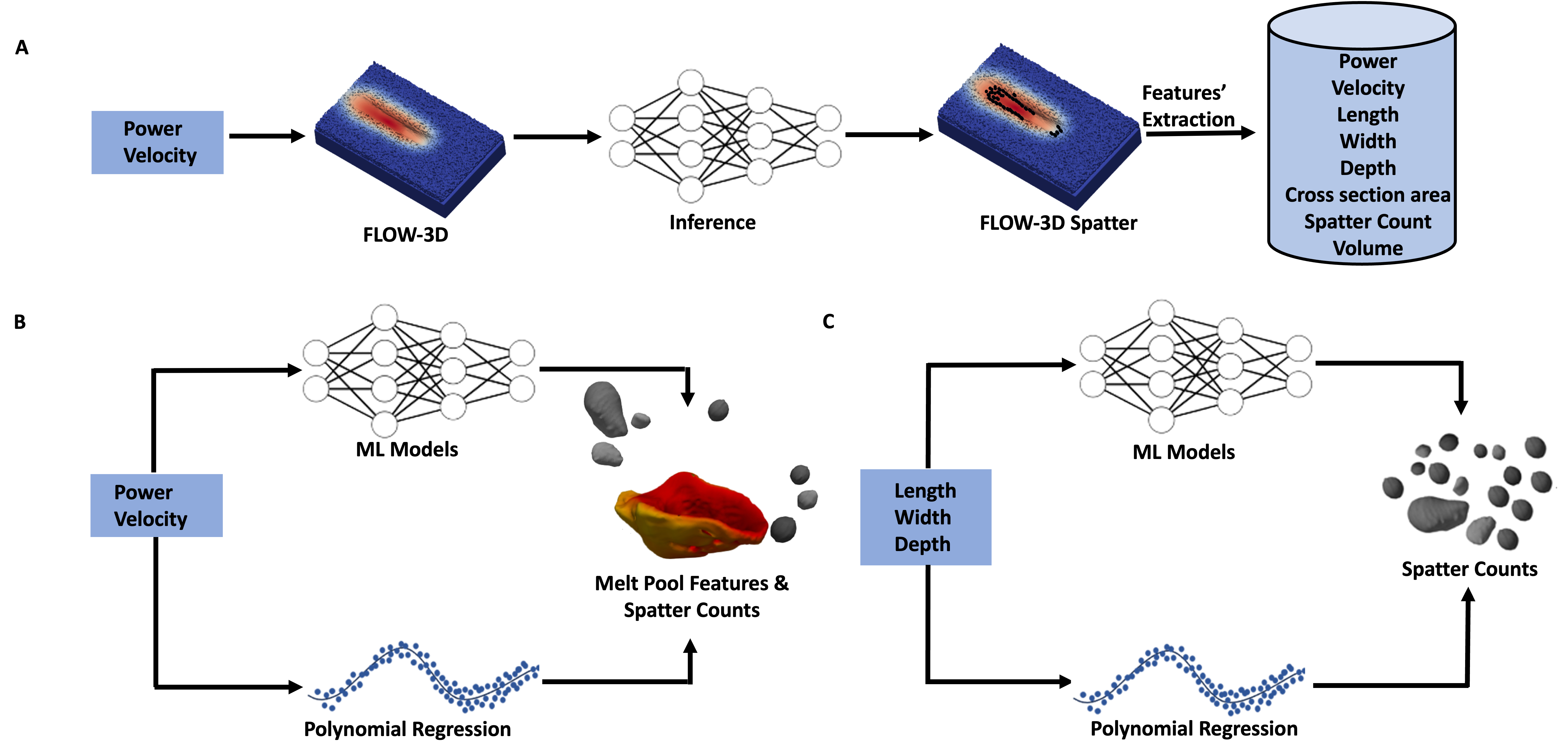}
\end{center}
    \caption{ Spatter dataset generated using OpenFOAM, a computationally expensive tool, was trained for classification task to differentiate between the two classes. Using the model as an inference, spatter count was predicted on a FLOW-3D, which is 18 times less computationally expensive tool. (a) Datasets consisting of process conditions, melt pool dimensions, geometry, and spatter count were collected from 281 FLOW-3D experiments (b) The process conditions was used as an input to either the ML model or polynomial regression to predict the melt pool dimensions, geometry features and the spatter count (c) The melt pool dimension was used as an input to either the ML model or polynomial regression to predict the spatter count}
    
\label{fig:description}
\end{figure*}

Building upon our previous investigations, this work employs both ML regression and symbolic regression to predict melt pool features measured at a stable state. These features include dimensions (length, width, and depth), geometric properties (cross-sectional area and volume), and the volume indicated by spatter. We collect the data set of the spatter count versus process conditions by augmenting OpenFOAM and FLOW-3D simulations via ML model developed for binary classification task for spatter/melt pool predictions. The motivation behind the techniques lies on the trade-off between the capability and computational costs of both simulation tools. FLOW-3D incorporates certain restrictive assumptions to reduce computational costs, which limit its ability to simulate spatter accurately. Specifically, it approximates the keyhole gas dynamics using estimates of recoil pressure and mass transfer, rather than directly simulating them. Furthermore, FLOW-3D does not account for the tangential surface tension component \cite{qu2022controlling, cheng2019computational}. In contrast, our OpenFOAM model directly simulates keyhole gas dynamics and incorporates tangential surface tension. The analysis indicated that the melt pool features can be predicted with a high R\(^2\) value from the process conditions (power and velocity). Similarly, the volume indicated as spatter can be predicted with a high R\(^2\)  value from either the process conditions or the dimensions of the melt pool, as shown in Fig. \ref{fig:description}. Our findings not only demonstrate the interconnection between process conditions, melt pool features, and spatter counts, but also provide an interpretable mathematical expression of the parameters derived from the input variables through symbolic regression models, which could be crucial for process optimization.
 
\section{Methods}
\subsection{LPBF Simulation Tools}
A CFD model was developed to analyze LPBF simulation using OpenFOAMv2012 using the icoReactingMultiphaseInterFoam (IRMIF) solver \cite{wirth2022implementation, lundkvist2019cfd, msmsac4a26bib17}. The IRMIF solver is based on the volume-of-fluid (VOF) method, where each phase is immiscible, and a clear boundary between each phase is calculated. The IRMIF solver handles fluid mechanics, including turbulent flow, laser beam sources with arbitrary beam shapes, heat transfer, and phase transitions such as solidification, melting, and evaporation. However, the solver did not account for the Marangoni effect due to the absence of the tangential component of the surface tension. The continuity and transient Navier stokes equation as described in equation (3) and (4) respectively.
 \vspace{-5mm}
\begin{equation}
\frac{\partial \rho}{\partial t}+\nabla \cdot(\bar{\rho} \vec{U})=0
\end{equation}

 \vspace{-10mm}
 
\begin{equation}
\frac{\partial(\rho \underline{u})}{\partial t}+\nabla(\rho \underline{u} \otimes \underline{u})=\nabla\left\{-p \cdot \underline{I}+\mu\left[\nabla \underline{u}+(\nabla \underline{u})^T\right]\right\}+\underline{F}
\end{equation}
where $\rho$ is the density, $\underline{u}$ is the fluid flow velocity vector, $t$ is the time, $p$ is the pressure, $I$ is the identity tensor, $\mu$  is the dynamic viscosity, and $\underline{F}$ is the volume force vector.
The energy equation in equation (5) was used to compute the temperature field.
\begin{equation}
\rho \frac{\delta E}{\delta t}=-\rho \vec{\nabla} \cdot \vec{u} E+\vec{\nabla} \cdot(\lambda \vec{\nabla} T+\overline{\bar{\tau}} \cdot \vec{u}) \pm S_h
\end{equation}
where $E$ is the mixed energy, $\lambda$ is the thermal conductivity. Further detail about the simulation can be found in our previous work \cite{ajenifujah2024integrating}.

FLOW-3D (v11.2) simulations are performed to accelerate process map development. FLOW-3D is a multiphysics simulation software produced by Flow Science, which provides more rapid estimations of the melt pool behavior than OpenFOAM. To create a dataset of FLOW-3D simulations, 281 SS316L single-track bare plate experiments are performed at varying processing parameters for a total length of 600 $\mu s$. During simulation, the FLOW-3D package solves the equations that describe mass transfer, momentum transfer, and energy transfer during the melting process. This simulation is carried out on a structured Cartesian mesh, with mesh elements sized at 10 $\mu m$. More specific information on the equations solved and the physical phenomena considered during the simulation can be found in \cite{ogoke2023convolutional, myers2023high, hemmasian2023surrogate, cook2020simulation, cheng2019computational}.

\subsection{Spatter Process Map Generation}
This study uses a range of computational techniques and machine learning methods to analyze OpenFOAM 3D simulation data and FLOW-3D to establish a spatter process map. Features such as position (x, y, z), velocity components ($v_x$, $v_y$, $v_z$), velocity magnitude, pressure, temperature, and density were extracted from the spatter and the melt pool following the methodology we previously outlined \cite{ajenifujah2024integrating}. These features of the melt pool and the spatter were passed into the ML model for the classification task. The prediction of spatter on the FLOW-3D dataset necessitates the extraction of similar features from the FLOW-3D as the OpenFOAM. The results of 281 different process conditions from the FLOW-3D dataset were pre-processed to extract the liquid fraction from the simulation as the meltpool. The velocity magnitude is calculated from the three velocity components ($v_x$, $v_y$, $v_z$) using Equation~(\ref{eq:velocity_magnitude}), producing scalar values that represent the speed at each point in the 3D grid.

\begin{equation}
U_{\text{magnitude}} = \sqrt{v_x^2 + v_y^2 + v_z^2}
\label{eq:velocity_magnitude}
\end{equation}

The features of the OpenFOAM dataset (velocity components ($v_x$, $v_y$, $v_z$), velocity magnitude, pressure, temperature, and density) are entered into the ML model for training and testing purposes. Before using the trained ML model as an inference for FLOW-3D datasets, the FLOW-3D dataset is pre-processed by aligning the range of each feature with OpenFOAM dataset. The FLOW-3D data set that represents one process condition is a representative of the average of the feature from the point the melt pool is determined to be stable to the end of the simulations. 

\subsection{Ensemble Learning and Polynomial Regression}
This study performs a regression task using the ML and polynomial regression models for the prediction of the features of the melting pool or the spatter as presented in Figure \ref{fig:description}. The following ML models are screened with the datasets: Extremely Randomized Trees (ExtraTrees), Extreme Gradient Boosting (XGB),  Random Forest (RF), Bagging, and k-Nearest Neighbors (KNN). The descriptions of the ML model are in Appendix A.1.1. To predict the spatter count,  two groups of features are identified, which are the process parameters and the dimensions of the melt pool. To predict the dimension and geometry of the melt pool, only the process parameters were implemented as input features of the model. These features are inputted into either the ML or polynomial regression models. Polynomial regression is particularly effective when linear models are insufficient to capture the complexity of the data.  In this case, the degree of the polynomial was vary in the range 2-6, allowing for higher-order interactions among the independent variables. A pipeline was constructed that combined polynomial feature transformation and linear regression. This pipeline streamlined the process of applying polynomial transformation and fitting the regression model. The Linear Regression model was trained on the transformed features, establishing the relationship between the independent and dependent features. Following model training, the coefficients and intercepts of the regression model were extracted. These values were used to formulate the polynomial equation, which expresses the relationship between the input variables and the target variable. The equation provides an explicit representation of the learned model, highlighting the contribution of each feature interaction to the prediction of the desired variable. The predictive performance of the model was evaluated using the coefficient of determination ($R^2$). The $R^2$ score reflects the proportion of variance in the dependent variable explained by the independent variables, with higher values indicating a better fit.

    



\section{Results and Discussion}
\subsection{Properties  variations and correlations across process conditions }
The melt pool dynamics in the LPBF process is shaped by a complex interplay of forces: gravity, Marangoni forces, buoyancy, and recoil pressure. Gravity exerts a uniform downward pull, influencing the overall shape and stability of the molten region. In contrast, Marangoni forces arise from surface tension gradients caused by temperature and compositional variations, driving fluid flow from hotter, lower-surface-tension regions to cooler, higher-surface-tension areas. These forces create convective currents that redistribute heat within the melt pool. Buoyancy, driven by density differences in the molten material due to temperature gradients, induces additional convective flows, with hotter, less dense material rising and cooler, denser material sinking. Together, gravity, Marangoni forces, and buoyancy determine the internal flow patterns and stability of the melt pool.

\begin{figure*}[htbp!]
\begin{center}
\includegraphics[width=1\linewidth]{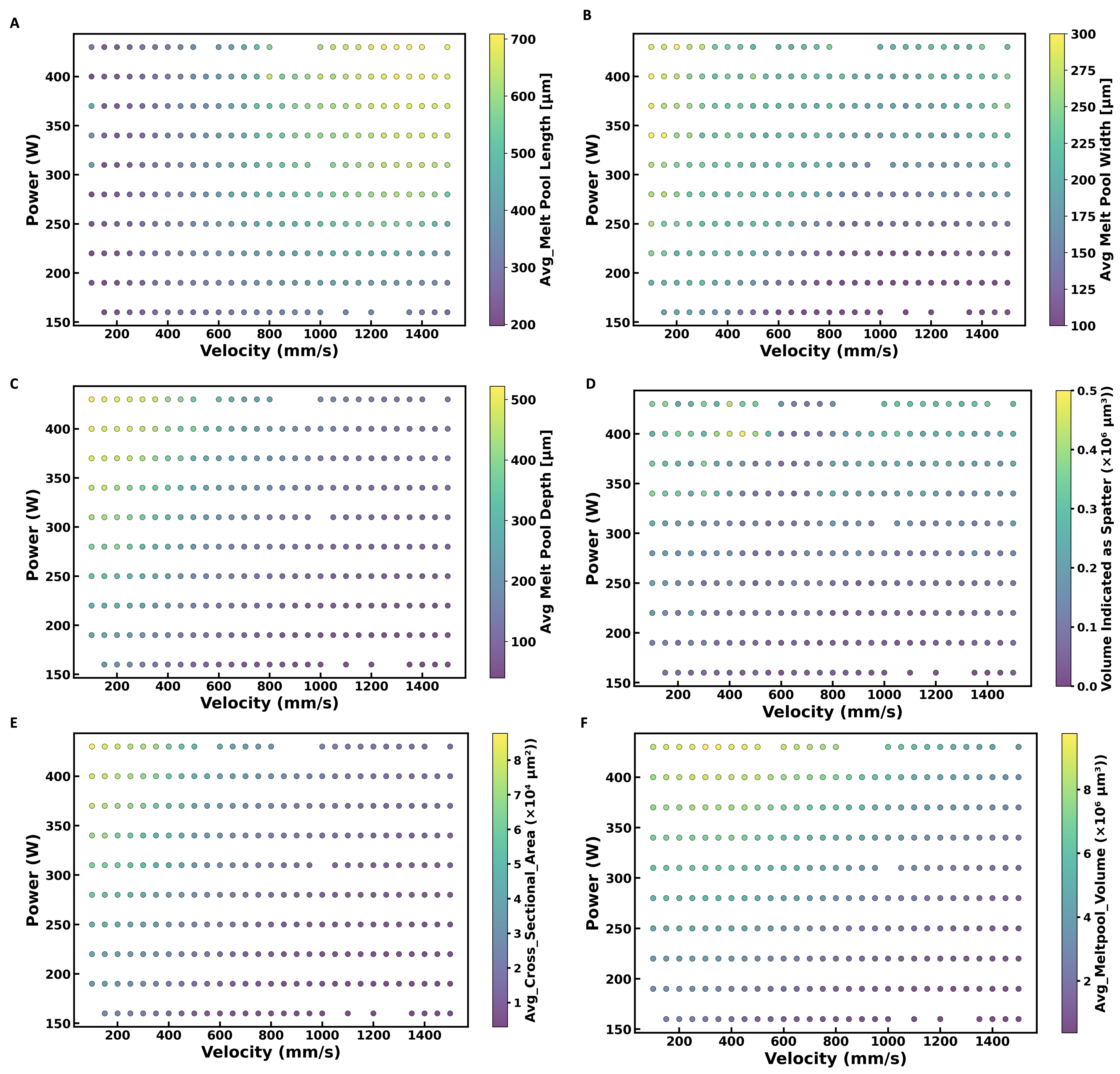}
\end{center}
    \caption{Distribution of the extracted parameters from the simulation of LPBF  (a) Melt pool length (b) Melt pool width (c) Melt pool depth (d) Volume indicated as spatter (e) Melt pool cross section area (f) Melt pool volume}
    
\label{fig:parameterdistribution}
\end{figure*}

The recoil pressure, generated by rapid vaporization at the surface of the melt pool, introduces localized pressure gradients that push the molten material away from the high-energy impact zone. This force influences the depth and shape of the melt pool and plays a significant role in material expulsion and spatter formation. The effect of these forces varies across four primary process regions: high-power low-speed, high-power high-speed, low-power low-speed, and low-power high-speed. In the low-speed, high-power region, recoil pressure and Marangoni forces often dominate, deforming the melt pool and potentially creating keyholes. Significant energy is concentrated in a small area for an extended period of time, producing a deep and wide melt pool with high volume and intense spatter, as shown in Figure \ref{fig:parameterdistribution} (b, c, d, e, f) due to strong recoil pressure and Marangoni-driven recirculation. In the high-power, high-speed region, the laser moves quickly over the material, resulting in a long, narrow melt pool with moderate depth, area, and volume, as shown in Figure \ref{fig:parameterdistribution}. Although the recoil pressure is high, the limited interaction time keeps the melt pool from becoming excessively deep or wide, and the generation of spatter is moderate.

\begin{figure*}[htbp!]
\begin{center}
\includegraphics[width=1\linewidth]{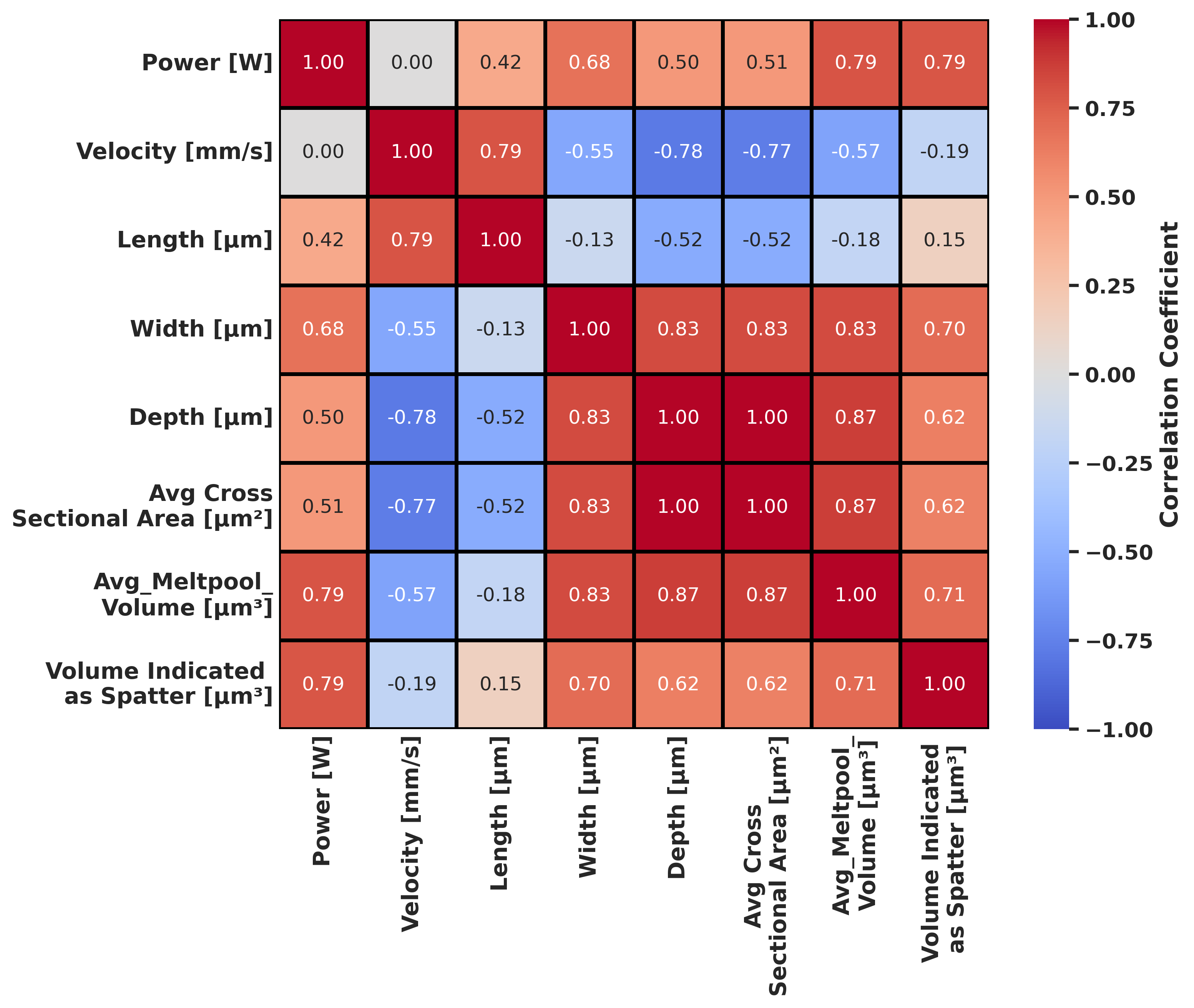}
\end{center}
    \caption{A correlation matrix displays the Pearson correlation coefficients between pairs of variables in the process conditions, melt pool dimensions and geometry }
    
\label{fig:correlation_matrix}
\end{figure*}

In the low-power low-speed region, the melt pool is shallow and broad with minimal spatter because a lower energy input produces less recoil pressure and vaporization. This configuration favors precision and surface quality over depth, making it suitable for applications where fine control is prioritized. Lastly, in the low-power high-speed region, the melt pool is very shallow and narrow, with negligible spatter because low-power and high-speed minimize melting. Gravity and buoyancy have a more pronounced effect in this region. Each of these regions produces distinct melt pool geometries, enabling manufacturers to tailor process parameters for specific design requirements, whether for high depth and volume or minimal surface impact.
\FloatBarrier
\FloatBarrier
 Figure \ref{fig:correlation_matrix} shows the correlation matrix that shows the Pearson correlation coefficients between pairs of variables, indicating the strength and direction of their linear relationships. The diagonal values are all 1.0, since each variable is perfectly correlated with itself. The color scale ranges from -1 (blue, indicating a strong negative correlation) to +1 (red, indicating a strong positive correlation). Positive correlations are shaded in red hues, while negative correlations appear in blue hues. Notable relationships include strong positive correlations between the width, depth, cross-sectional area, and volume of the melt pool, with coefficients close to or above 0.8, indicating that as one of these variables increases, the corresponding pairing variable increases as well. Furthermore, the power and volume indicated as spatter show a high correlation (0.79), suggesting that higher power levels may lead to higher volumes of spatter. Moderate positive correlations are observed between power and width (0.68), power and depth (0.50), and power and avg melt pool volume (0.79), which implies that an increase in power generally corresponds to increases in these variables. In contrast, velocity exhibits negative correlations with depth (-0.78), cross-sectional area of avg (-0.77) and volume of the avg melt pool (-0.57), indicating that higher velocities are associated with smaller depths, cross-sectional areas, and volumes of the melt pool. Weak correlations were observed in the following combination of variables: Length and width (-0.13), length and avalanche meltpool volume (-0.18), velocity and volume indicated as splinter (-0.19), and length and volume indicated as splinter (0.15). These low correlation values imply that these pairs of variables have minimal linear relationships, which means that changes in one variable do not consistently predict changes in the other.

\FloatBarrier
\FloatBarrier

\subsection{Machine Learning regression prediction}
The performance of five machine learning algorithms: Random forests (RF), extra trees (ExtraTree), bagging, k-nearest neighbors (KNN), and gradient boost (GB) were evaluated using input features derived from process conditions and melt pool dimensions. With process conditions (power and velocity) as input, the models can predict the dimensions of the melt pool (length, width, and depth), geometric features (area and volume) and the volume indicated as spatter as shown in Figure \ref{fig:description}b. Furthermore, using the dimensions of the melting pool (length, width, and depth) as input, the models can predict the volume indicated as a spatter, as shown in Figure \ref{fig:description}c. The hyperparameters implemented for the ML models, such as the number of estimators, the maximum depth, the number of neighbors, and the learning rates, are listed in Table 1. Table 2 summarizes the performance metrics used, including R\(^2\) scores for both training and testing sets, as well as mean absolute error (MAE) for both sets. The R\(^2\) score measures the proportion of variance in the dependent variable that is predictable from the independent variables, while the MAE quantifies the average magnitude of the prediction errors, regardless of their direction.
 Using power and velocity as the model input, the R\(^2\) predictions for the melt pool length data set show strong performance, with the training data set R\(^2\) values exceeding 98\%, and the lowest R\(^2\) for the test dataset reaching 95.6\% when using the ExtraTree model. Similarly, high R\(^2\) values were observed for depth, melt pool area, and volume, with R\(^2\) scores for training and test sets exceeding 97\%. For depth, R\(^2\) values remained consistently high ($>$ 97\%) on both the training and the test data. However, a slight drop in the R\(^2\) prediction was observed for width, with the lowest values appearing in the KNN model, where the training and test R\(^2\) scores were 97.4\% and 91\%, respectively. A further drop in the prediction of the volume indicated as spatter; in this case, the most promising model was RF, with training and test R\(^2\) values of 94.1\% and 83.4\%, respectively.

Introducing logarithmic terms to the input was considered a strategy to improve accuracy, as it can better capture nonlinear relationships between the variables. Upon adjusting the input of the model from power and velocity to power, velocity and log(Velocity), the ExtraTree model demonstrated the most significant improvement in test accuracy. The test accuracy of ExtraTree increased from 79.6\% to 84.1\%, while its training accuracy improved from 91.4\% to 94.3\%. This enhancement highlights the potential of input manipulation through logarithmic transformations to improve prediction accuracy by effectively modeling nonlinear dependencies.

 Because power and velocity are fixed during a particular process operation, our results demonstrate that these process conditions can effectively predict the characteristics of a stable melt pool. When using the dimensions of the melt pool (length, width and depth) as inputs into the ML models to predict the volume indicated as a spatter, we observed an accuracy comparable to when power and velocity were used as inputs. The KNN model achieved the highest test accuracy at 82.3\%, with a training accuracy of 85.8\%. Although the Extra Tree model had the highest training accuracy at 90.5\%, its test accuracy was slightly lower at 80.1\%. Upon introducing logarithmic terms to the model input, changing them from length, width, and depth to log length, width, depth, log width, and log depth, the most significant improvement was observed in the Extra Tree model, with training and test accuracies reaching 95.8\% and 87.5\%, respectively. This improvement highlights the effectiveness of incorporating logarithmic terms to capture the nonlinear relationships within melt-pool dimensions, enhancing the prediction of volume indicated as spatter.
  
\begin{table}[ht]
\vspace{-2cm}

\caption{The hyperparameters used to train different ML models with varying inputs and outputs.
}
\centering
\hspace*{-1cm}
%
\begin{adjustbox}{width=\textheight,angle=90} 

\footnotesize
\cellcolor{white}
\color{black}
\begin{tabular}{|c|c|c|c|c|c|c|}
\hline
\textbf{Model} & \textbf{Input} & \textbf{Output} & \textbf{n\_estimators} & \textbf{max\_depth} & \textbf{n\_neighbors} & \textbf{LR} \\ \hline

RF & [Power, Velocity] & [Length] & 5 & 8 & & \\ \hline
ExtraTree & [Power, Velocity] & [Length] & 5 & 7 & & \\ \hline
Bagging & [Power, Velocity] & [Length] & 3 & 7 & & \\ \hline
KNN & [Power, Velocity] & [Length] & & & 2 & \\ \hline
GB & [Power, Velocity] & [Length] & 70 & 6 & & 0.1 \\ \hline
RF & [Power, Velocity] & [Width] & 6 & 7 & & \\ \hline
ExtraTree & [Power, Velocity] & [Width] & 8 & 8 & & \\ \hline
Bagging & [Power, Velocity] & [Width] & 5 & 7 & & \\ \hline
KNN & [Power, Velocity] & [Width] & & & 3 & \\ \hline
GB & [Power, Velocity] & [Width] & 80 & 4 & & 0.1 \\ \hline
RF & [Power, Velocity] & [Depth] & 8 & 6 & & \\ \hline
ExtraTree & [Power, Velocity] & [Depth] & 5 & 7 & & \\ \hline
Bagging & [Power, Velocity] & [Depth] & 3 & 7 & & \\ \hline
KNN & [Power, Velocity] & [Depth] & & & 4 & \\ \hline
GB & [Power, Velocity] & [Depth] & 80 & 5 & & 0.1 \\ \hline
RF & [Power, Velocity] & [Avg\_Cross\_Sectional\_Area] & 8 & 6 & & \\ \hline
ExtraTree & [Power, Velocity] & [Avg\_Cross\_Sectional\_Area] & 11 & 8 & & \\ \hline
Bagging & [Power, Velocity] & [Avg\_Cross\_Sectional\_Area] & 4 & 7 & & \\ \hline
KNN & [Power, Velocity] & [Avg\_Cross\_Sectional\_Area] & & & 3 & \\ \hline
GB & [Power, Velocity] & [Avg\_Cross\_Sectional\_Area] & 4 & 9 & & \\ \hline
RF & [Power, Velocity] & [Avg\_Meltpool\_Volume] & 8 & 9 & & \\ \hline
ExtraTree & [Power, Velocity] & [Avg\_Meltpool\_Volume] & 11 & 8 & & \\ \hline
Bagging & [Power, Velocity] & [Avg\_Meltpool\_Volume] & 4 & 9 & & \\ \hline
KNN & [Power, Velocity] & [Avg\_Meltpool\_Volume] & & & 4 & \\ \hline
GB & [Power, Velocity] & [Avg\_Meltpool\_Volume] & 4 & 9 & & \\ \hline
RF & [Power, Velocity] & [Volume Indicated as Spatter] & 6 & 5 & & \\ \hline
ExtraTree & [Power, Velocity] & [Volume Indicated as Spatter] & 10 & 6 & & \\ \hline
Bagging & [Power, Velocity] & [Volume Indicated as Spatter] & 6 & 4 & & \\ \hline
KNN & [Power, Velocity] & [Volume Indicated as Spatter] & & & 5 & \\ \hline
GB & [Power, Velocity] & [Volume Indicated as Spatter] & 23 & 4 & & 0.1 \\ \hline
RF & [Power, Velocity, log\_Velocity] & [Volume Indicated as Spatter] & 12 & 4 & & \\ \hline
ExtraTree & [Power, Velocity, log\_Velocity] & [Volume Indicated as Spatter] & 2 & 6 & & \\ \hline
Bagging & [Power, Velocity, log\_Velocity] & [Volume Indicated as Spatter] & 4 & 7 & & \\ \hline
KNN & [Power, Velocity, log\_Velocity] & [Volume Indicated as Spatter] & & & 5 & \\ \hline
GB & [Power, Velocity, log\_Velocity] & [Volume Indicated as Spatter] & 21 & 4 & & \\ \hline
RF & [Length, Width, Depth] & [Volume Indicated as Spatter] & 6 & 4 & & \\ \hline
ExtraTree & [Length, Width, Depth] & [Volume Indicated as Spatter] & 30 & 6 & & \\ \hline
Bagging & [Length, Width, Depth] & [Volume Indicated as Spatter] & 70 & 4 & & \\ \hline
KNN & [Length, Width, Depth] & [Volume Indicated as Spatter] & & & 9 & \\ \hline
GB & [Length, Width, Depth] & [Volume Indicated as Spatter] & 35 & 2 & & 0.1 \\ \hline
RF & [log\_Length, Width, Depth, log\_Width, log\_Depth] & [Volume Indicated as Spatter] & 6 & 4 & & \\ \hline
ExtraTree & [log\_Length, Width, Depth, log\_Width, log\_Depth] & [Volume Indicated as Spatter] & 100 & 7 & & \\ \hline
Bagging & [log\_Length, Width, Depth, log\_Width, log\_Depth] & [Volume Indicated as Spatter] & 7 & 5 & & \\ \hline
KNN & [log\_Length, Width, Depth, log\_Width, log\_Depth] & [Volume Indicated as Spatter] & & & 8 & \\ \hline
GB & [log\_Length, Width, Depth, log\_Width, log\_Depth] & [Volume Indicated as Spatter] & 35 & 3 & & \\ \hline

\end{tabular}

\end{adjustbox}

\end{table}

\FloatBarrier

\FloatBarrier
\begin{table}[ht]
\vspace{-2cm}
\caption{ Performance evaluation of the ML models using the $R^2$ metric and mean absolute error (MAE) for both training and test datasets.}
\centering
\hspace*{-1cm}
%
\begin{adjustbox}{width=\textheight,angle=90} 

\footnotesize
\cellcolor{white}
\color{black}
\begin{tabular}{|c|c|c|c|c|c|c|}
\hline
\textbf{Model} & \textbf{Input} & \textbf{Output} & \textbf{R\(^2\)(Train)} & \textbf{R\(^2\)(Test)} & \textbf{Train MAE} & \textbf{Test MAE}  \\ \hline

RF & [Power, Velocity] & [Length] & 0.9955 & 0.9611 & 6.1791 & 16.0696   \\ \hline
ExtraTree & [Power, Velocity] & [Length] & 0.9876 & 0.9557 & 10.4751 & 17.7457  \\ \hline
Bagging & [Power, Velocity] & [Length] & 0.9935 & 0.966 & 7.1818 & 15.0308 \\ \hline
KNN & [Power, Velocity] & [Length] & 0.9833 & 0.9681 & 14.1846 & 15.3929  \\ \hline
GB & [Power, Velocity] & [Length] & 0.9999 & 0.9598 & 1.0422 & 14.6793  \\ \hline
RF & [Power, Velocity] & [Width] & 0.983 & 0.9319 & 4.8998 & 9.9063  \\ \hline
ExtraTree & [Power, Velocity] & [Width] & 0.9899 & 0.9425 & 3.7299 & 8.6522  \\ \hline
Bagging & [Power, Velocity] & [Width] & 0.9824 & 0.9355 & 4.945 & 9.729  \\ \hline
KNN & [Power, Velocity] & [Width] & 0.9741 & 0.9109 & 6.2479 & 10.7778 \\ \hline
GB & [Power, Velocity] & [Width] & 0.9913 & 0.9374 & 3.6616 & 9.5184 \\ \hline
RF & [Power, Velocity] & [Depth] & 0.9948 & 0.9799 & 6.5786 & 12.8118 \\ \hline
ExtraTree & [Power, Velocity] & [Depth] & 0.9928 & 0.9889 & 7.0481 & 9.8546  \\ \hline
Bagging & [Power, Velocity] & [Depth] & 0.9932 & 0.9755 & 6.4861 & 14.3183  \\ \hline
KNN & [Power, Velocity] & [Depth] & 0.9935 & 0.9766 & 6.8679 & 12.1577  \\ \hline
GB & [Power, Velocity] & [Depth] & 0.9998 & 0.9868 & 1.4649 & 10.5286  \\ \hline
RF & [Power, Velocity] & [Avg\_Cross\_Sectional\_Area] & 0.9968 & 0.988 & 662.3621 & 1438.8822  \\ \hline
ExtraTree & [Power, Velocity] & [Avg\_Cross\_Sectional\_Area] & 0.9984 & 0.9913 & 534.358 & 1325.6511 \\ \hline
Bagging & [Power, Velocity] & [Avg\_Cross\_Sectional\_Area] & 0.995 & 0.9844 & 891.3867 & 1712.9828  \\ \hline
KNN & [Power, Velocity] & [Avg\_Cross\_Sectional\_Area] & 0.9948 & 0.9821 & 907.0427 & 1771.4683 \\ \hline
GB & [Power, Velocity] & [Avg\_Cross\_Sectional\_Area] & 0.9954 & 0.9835 & 821.5204 & 1765.4271  \\ \hline
RF & [Power, Velocity] & [Avg\_Meltpool\_Volume] & 0.9983 & 0.9911 & 75125.5827 & 179674.6205 \\ \hline
ExtraTree & [Power, Velocity] & [Avg\_Meltpool\_Volume] & 0.999 & 0.9932 & 57304.9607 & 154726.1848  \\ \hline
Bagging & [Power, Velocity] & [Avg\_Meltpool\_Volume] & 0.9977 & 0.9903 & 80251.047 & 187474.9603 \\ \hline
KNN & [Power, Velocity] & [Avg\_Meltpool\_Volume] & 0.9909 & 0.9679 & 174572.6923 & 263140.1786  \\ \hline
GB & [Power, Velocity] & [Avg\_Meltpool\_Volume] & 0.9977 & 0.9903 & 80251.047 & 187474.9603  \\ \hline
RF & ['Power', 'Velocity'] & ['Volume Indicated as Spatter']  & 0.9414 & 0.8337 & 18709.6168 & 28853.8388  \\ \hline
ExtraTree & [Power, Velocity] & [Volume Indicated as Spatter] & 0.8918 & 0.796 & 22430.2771 & 31858.8664  \\ \hline
Bagging & [Power, Velocity] & [Volume Indicated as Spatter] & 0.9135 & 0.8206 & 23017.1701 & 30280.1787  \\ \hline
KNN & [Power, Velocity] & [Volume Indicated as Spatter] & 0.9057 & 0.8312 & 21643.8462 & 29152.9762 \\ \hline
GB & [Power, Velocity] & [Volume Indicated as Spatter] & 0.9352 & 0.8292 & 20081.6918 & 29010.1023  \\ \hline
RF & [Power, Velocity, log\_Velocity] & [Volume Indicated as Spatter] & 0.9095 & 0.8006 & 23347.2729 & 30977.7348 \\ \hline
ExtraTree & [Power, Velocity, log\_Velocity] & [Volume Indicated as Spatter] & 0.9435 & 0.8411 & 18106.0869 & 27507.3048 \\ \hline
Bagging & [Power, Velocity, log\_Velocity] & [Volume Indicated as Spatter] & 0.9545 & 0.8399 & 14749.7506 & 27972.8045  \\ \hline
KNN & [Power, Velocity, log\_Velocity] & [Volume Indicated as Spatter] & 0.9059 & 0.8159 & 21540.5128 & 30233.9286 \\ \hline
GB & [Power, Velocity, log\_Velocity] & [Volume Indicated as Spatter] & 0.9252 & 0.8222 & 21730.3966 & 29863.3115  \\ \hline
RF & [Length, Width, Depth] & [Volume Indicated as Spatter] & 0.8698 & 0.7563 & 26654.2475 & 32366.5143  \\ \hline
ExtraTree & [Length, Width, Depth] & [Volume Indicated as Spatter] & 0.9045 & 0.801 & 23358.7838 & 30081.3716 \\ \hline
Bagging & [Length, Width, Depth] & [Volume Indicated as Spatter] & 0.8701 & 0.7624 & 26556.1096 & 31717.2  \\ \hline
KNN & [Length, Width, Depth] & [Volume Indicated as Spatter] & 0.8581 & 0.8233 & 27269.6581 & 27395.172  \\ \hline
GB & [Length, Width, Depth] & [Volume Indicated as Spatter] & 0.8559 & 0.7554 & 29410.0416 & 34711.6878\\ \hline
RF & [log\_Length, Width, Depth, log\_Width, log\_Depth] & [Volume Indicated as Spatter] & 0.9033 & 0.8349 & 0.1924 & 0.2562\\ \hline
ExtraTree & [log\_Length, Width, Depth, log\_Width, log\_Depth] & [Volume Indicated as Spatter] & 0.9677 & 0.8753 & 0.1112 & 0.2173\\ \hline
Bagging & [log\_Length, Width, Depth, log\_Width, log\_Depth] & [Volume Indicated as Spatter] & 0.9263 & 0.831 & 0.174 & 0.2508\\ \hline
KNN & [log\_Length, Width, Depth, log\_Width, log\_Depth] & [Volume Indicated as Spatter] & 0.7711 & 0.7508 & 0.3002 & 0.2856 \\ \hline
GB & [log\_Length, Width, Depth, log\_Width, log\_Depth] & [Volume Indicated as Spatter] & 0.9388 & 0.842 & 0.1597 & 0.2284  \\ \hline

\end{tabular}
\end{adjustbox}
\end{table}

\FloatBarrier
\subsection{Polynomial regression prediction}
\begin{figure*}[htbp!]
\begin{center}
\includegraphics[width=1\linewidth]{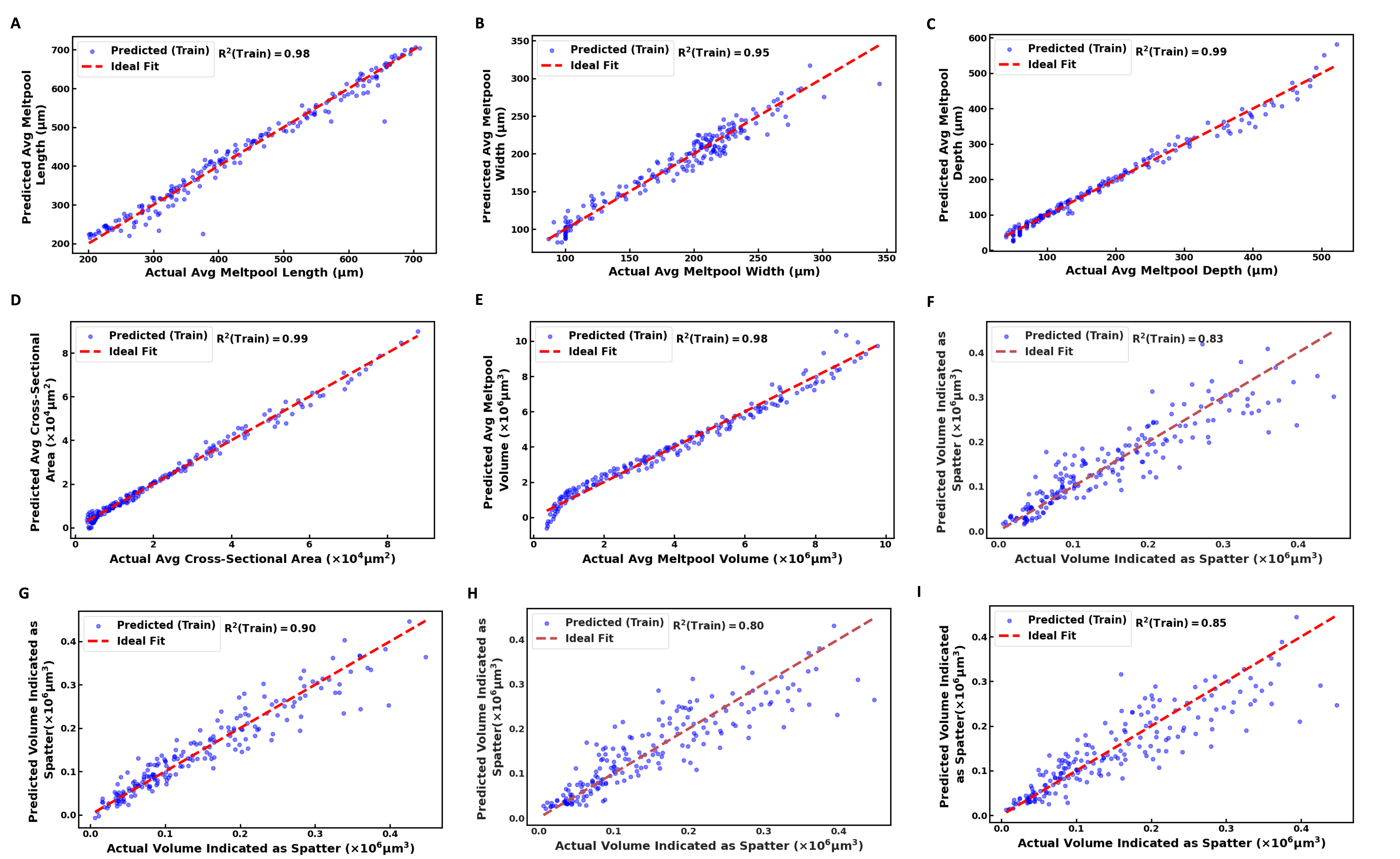}
\end{center}
    \caption{Polynomial regression fitting on the training dataset using either process conditions or melt pool dimension as an input into the model to predict the following (A) Melt pool length prediction using power and velocity as the model input (B) Melt pool width prediction using power and velocity as the model input (C) Melt pool depth prediction using power and velocity as the model input (D)  Melt pool top cross sectional area prediction using power and velocity as the model input(E)  Melt pool volume prediction using power and velocity as the model input (F) Prediction of volume indicated as spatter  using power and velocity as the model input (G) Prediction of volume indicated as spatter  using power and velocity as the model input Prediction of volume indicated as spatter  using Power, Velocity, and $\log$(Velocity) as the model input (H) Prediction of volume indicated as spatter  using length, width and depth as the model input (I) Prediction of volume indicated as spatter  using $\log$(length), width, depth, $\log$(width), and $\log$(depth) as the model input }
    
\label{fig:regression_training}
\end{figure*}

\begin{figure*}[htbp!]
\begin{center}
\includegraphics[width=1\linewidth]{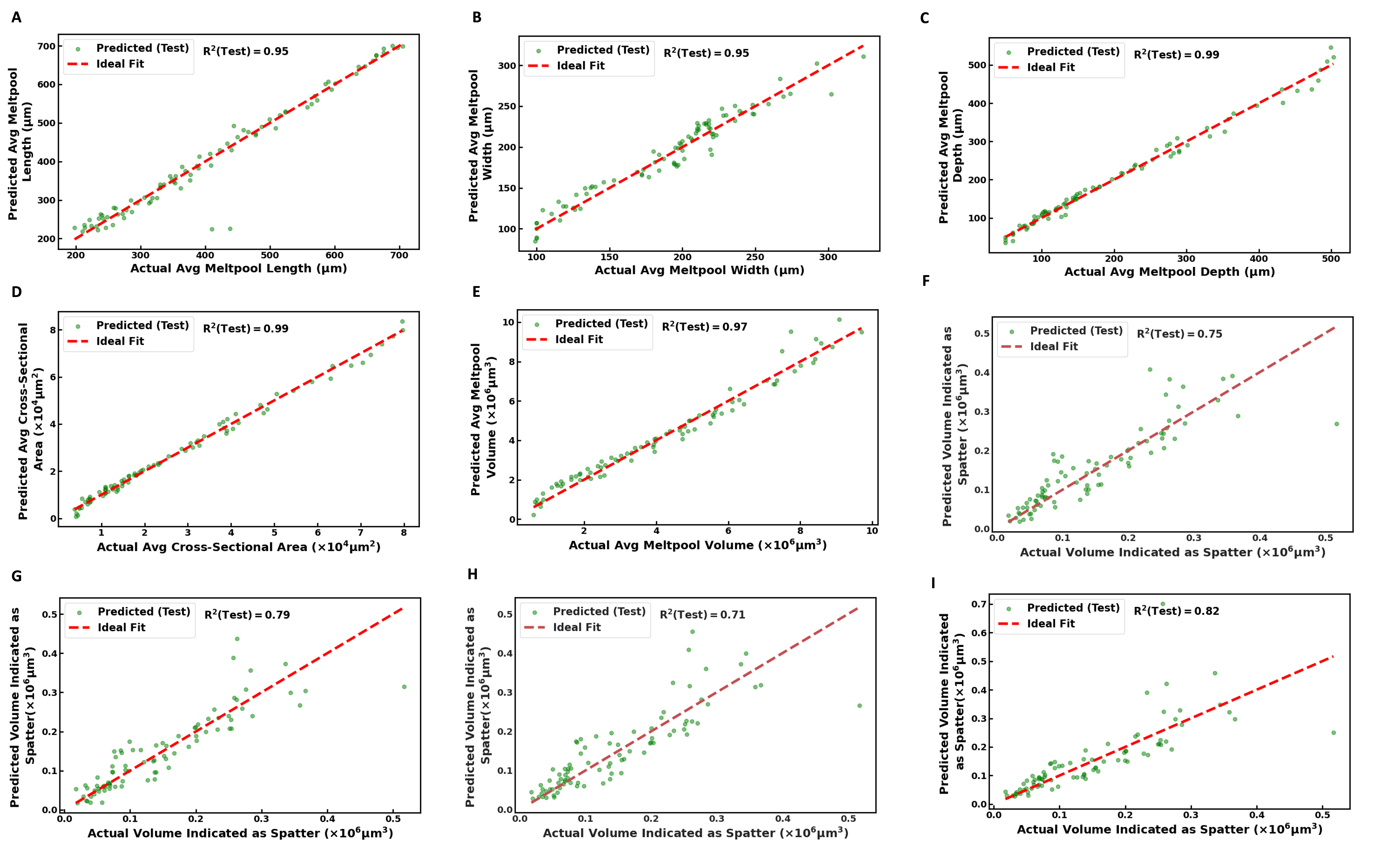}
\end{center}
    \caption{Polynomial regression fitting for the test dataset (A) Melt pool length prediction using power and velocity as the model input (B) Melt pool width prediction using power and velocity as the model input (C) Melt pool depth prediction using power and velocity as the model input (D)  Melt pool top cross sectional area prediction using power and velocity as the model input(E)  Melt pool volume prediction using power and velocity as the model input (F) Prediction of volume indicated as spatter  using power and velocity as the model input (G) Prediction of volume indicated as spatter  using power and velocity as the model input Prediction of volume indicated as spatter  using Power, Velocity, and $\log$(Velocity) as the model input (H) Prediction of volume indicated as spatter  using length, width and depth as the model input (I) Prediction of volume indicated as spatter  using $\log$(length), width, depth, $\log$(width), and $\log$(depth) as the model input}
    
\label{fig:regression_testing}
\end{figure*}

To understand the relationship between the input and the output, polynomial regression was implemented, as it offers interpretability by breaking down the influence of each input variable and its interactions on the output prediction. This interpretability arises from the polynomial terms in the regression equation, where each term directly reflects the contribution of an input variable or a combination of variables to the overall prediction. Upon conducting polynomial fitting, where process conditions (power and velocity) were input while any of the melt pool dimension or geometry features served as the output, high performance was observed with training and testing R\(^2\) exceeding 94\%, as shown in Figures \ref{fig:regression_training} and \ref{fig:regression_testing}, respectively, and their values listed in Table 3 with the corresponding degree of polynomial implemented. The lowest performance among the melt pool dimension and geometry features was with respect to width (training $R^2$: 95\% and testing $R^2$: 95\%). Achieving the reported performance for width required using a higher-order polynomial (degree 5), indicating complexity in calculating the width as more interacting terms are involved.

A polynomial regression model was developed to predict the volume indicated as a spatter using four separate input features of which two sets of the process condition and the other two sets from melt pool dimensions: [Power, Velocity, and $\log$(Velocity)] and [$\log$(length), width, depth, $\log$(width), and $\log$(depth)], respectively. Logarithmic transformations were applied to the input process conditions and melt pool dimensions to enhance spatter prediction, as the logarithmic transformation can stabilize the variance across different ranges of the data. In addition, logarithmic transformations reduce skewness in variables. An improvement in both the training and the test accuracy was observed when a logarithmic transformation was applied to both the process condition and the input of the dimension of the melt pool. With respect to the process condition, we achieved increments of 8.4\% and 5.3\% in the training and testing R\(^2\) values, respectively, by applying the logarithmic terms. A sixth-degree polynomial model was identified as the best fit, indicating the need to express complex non-linear interactions between these features in the process conditions. Concurrently, an improvement in performance was observed when a logarithmic transformation was applied to the melt pool dimensions as input into the polynomial regression model. Training and testing R\(^2\) increased by 6.3\% and 15.5\%, respectively, after the introduction of logarithmic terms.

\begin{figure*}[htbp!]
\begin{center}
\includegraphics[width=1.1\linewidth]{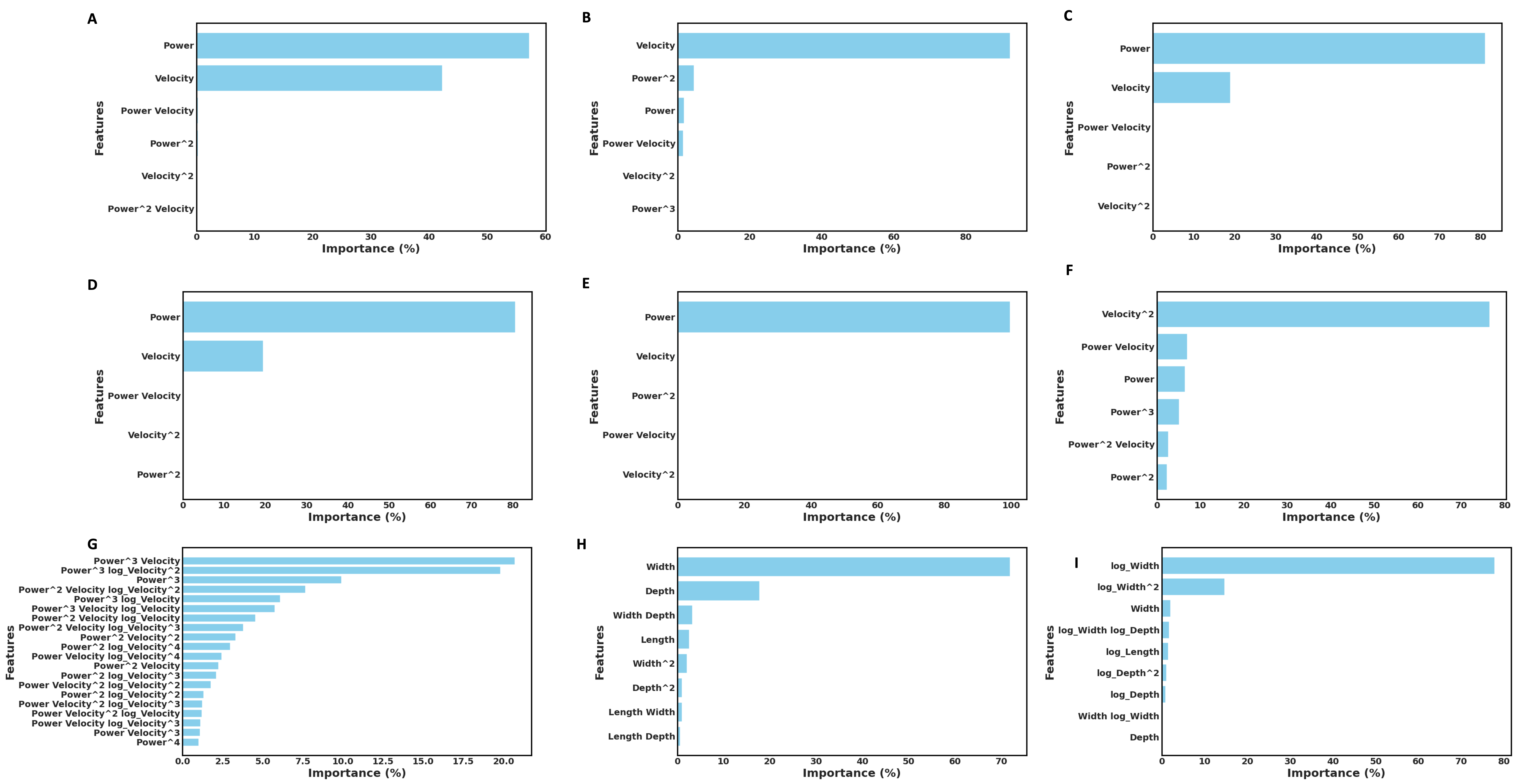}
\end{center}
    \caption{ Display of the top feature importances from the derived equation in polynomial regression  (A) Melt pool length prediction using power and velocity as the model input (B) Melt pool width prediction using power and velocity as the model input (C) Melt pool depth prediction using power and velocity as the model input (D)  Melt pool top cross sectional area prediction using power and velocity as the model input(E)  Melt pool volume prediction using power and velocity as the model input (F) Prediction of volume indicated as spatter  using power and velocity as the model input (G) Prediction of volume indicated as spatter  using power and velocity as the model input Prediction of volume indicated as spatter  using Power, Velocity, and $\log$(Velocity) as the model input (H) Prediction of volume indicated as spatter  using length, width and depth as the model input (I) Prediction of volume indicated as spatter  using $\log$(length), width, depth, $\log$(width), and $\log$(depth) as the model input}
    
\label{fig:Importance_PR}
\end{figure*}

The equations for different trained features are listed for the properties of the melt pool in Table 4, for the volume indicated as spatter in Tables 5 and Table A.6 in the Appendix. To quantify the importance of features in the equations, the absolute coefficients of the top contributing variables were extracted and their percentage contributions are shown in Figure \ref{fig:Importance_PR}. The generated equations indicate that the prediction of variables like length and depth depends more on the introduced power than on the velocity, while the width depends more on the velocity than on the power.
 At a particular velocity, the depth or length value will intuitively depend on the applied power, since the power influences the dynamics of the melt pool, the amount of liquid fraction and the rate of solidification. However, the fact that velocity has a stronger importance than power for the width suggests a low sensitivity of power over a wide range of velocities, as the melt pool motion and temperature gradient are influenced, making the stabilization of the width more dependent on velocity. The greater magnitude of length and depth range compared to width range, as shown in Figure \ref{fig:parameterdistribution} in the investigated process conditions, could have led to the strong importance of power in the geometry features of the melt pool (cross-sectional area and volume).
Since the influence of power is more dominant in the equations for both length and depth, the dependency of the volume of the melt pool on these variables is likely influenced by power, as observed in Figure \ref{fig:Importance_PR}E. The effect of width, being more influenced by velocity, arises from the cross-sectional area but is not strong enough to dominate over the power. The term $\text{Velocity}^2$ is the most important variable for the predicted volume indicated as spatter when the process condition was used as the input of the model, as shown in Figure \ref{fig:Importance_PR}F. Upon applying a logarithmic transformation to the process condition in Figure \ref{fig:Importance_PR}G, it was found that the top two ranking variables are "$\text{Power}^3$, $\text{Velocity}$" and "$\text{Power}^3$, $\log(\text{Velocity})$", indicating that they are highly influential in determining the volume indicated as a spatter when a logarithmic transformation is applied to the process conditions. 
 In the case of applying the dimensions of the melt pool as input to the regression model, it was found that the width followed by the depth is the most important, as shown in Figure \ref{fig:Importance_PR}H. Upon applying a logarithmic transformation to the dimension of the melt pool in Figure \ref{fig:Importance_PR} I, the width derivative in the form of $\log(\text{width})$ and $\log(\text{width})^2$ was found to be the first two variables to be most significant, indicating the prominent role of width before and after applying the logarithmic transformation to the prediction of volume indicated as spatter. 
      

\FloatBarrier
\hspace*{-20mm}
\begin{table}[ht]

\centering
\caption{Performance of  polynomial regression model with the degree of the polynomial in predicting melt pool dimensions, geometry and volume indicated as spatter}


\footnotesize
\begin{tabular}{|c|c|c|c|c|c|c|c|c|c|c|}

\hline
\textbf{Input} & \textbf{Predicted Features} & \textbf{Order} & \textbf{R\(^2\)(Train)} & \textbf{R\(^2\)(Test)} \\ 
\hline
[Power, Velocity] & Avg Meltpool Length & 3 & 0.98 & 0.95 \\
\hline
[Power, Velocity] & Avg Meltpool Width & 5 & 0.95 & 0.95 \\
\hline
[Power, Velocity] & Avg Meltpool Depth & 2 & 0.99 & 0.99 \\
\hline
[Power, Velocity] & Avg Cross Sectional Area & 2 & 0.99 & 0.99 \\
\hline
[Power, Velocity] & Avg Meltpool Volume & 2 & 0.98 & 0.97 \\
\hline
[Power, Velocity] & Volume Indicated as Spatter & 6 & 0.83 & 0.75 \\
\hline
[Power, Velocity, log Velocity] & Volume Indicated as Spatter & 6 & 0.9 & 0.79 \\
\hline
[Length, Width, Depth] & Volume Indicated as Spatter & 3 & 0.8 & 0.71 \\
\hline
[log Length, Width, Depth, log Width, log Depth] & Volume Indicated as Spatter & 2 & 0.85 & 0.82 \\
\hline
\end{tabular}


\end{table}
\FloatBarrier

\begin{table}[ht]
\centering
\caption{Polynomial regression equations for melt pool dimensions and geometry features}
\renewcommand{\arraystretch}{1.2}
\begin{tabular}{|c|>{\raggedright\arraybackslash}p{12cm}|c|}
\hline
\textbf{Features} & \textbf{Equation} & \(\mathbf{R^2(Train)}\) \\
\hline
Length & 
$170.3876 + 0.7513P - 0.5552V - 0.0032P^2 + 0.0034PV + 5.40 \times 10^{-4}V^2 + 4.19 \times 10^{-6}P^3 - 4.83 \times 10^{-6}P^2V + 4.49 \times 10^{-7}PV^2 - 3.46 \times 10^{-7}V^3$ & 0.98 \\
\hline
Width & 
$14.7778  - 0.0065P + 0.3516V + 0.0173P^2 - 0.0056PV + 1.94 \times 10^{-4}V^2 - 9.14 \times 10^{-5}P^3 + 3.31 \times 10^{-5}P^2V - 4.99 \times 10^{-6}PV^2 - 1.35 \times 10^{-7}V^3 + 1.96 \times 10^{-7}P^4 - 1.12 \times 10^{-7}P^3V + 3.67 \times 10^{-8}P^2V^2 - 4.82 \times 10^{-9}PV^3 + 7.74 \times 10^{-10}V^4$ & 0.95 \\
\hline
Depth & 
$53.7694 + 1.5055P - 0.3504V - 2.92 \times 10^{-4}P^2 - 7.54 \times 10^{-4}PV + 2.12 \times 10^{-4}V^2$ & 0.99 \\
\hline
Area & 
$4176.5581  + 224.9810P - 54.3024V - 0.0011P^2 - 0.1333PV + 0.0353V^2$ & 0.99 \\
\hline
Volume & 
$-1262141.6793  + 21113.6671P + 7.5091V + 17.3061P^2 - 9.5400PV - 0.4026V^2$ & 0.98 \\
\hline
\end{tabular}
\end{table}
\begin{table}[ht]
\centering
\caption{Polynomial regression equations for melt pool volume indicated as spatter, as the equations are derived from the model input}
\renewcommand{\arraystretch}{1.2}
\begin{tabular}{|c|>{\raggedright\arraybackslash}p{12cm}|c|}
\hline
\textbf{Features} & \textbf{Equation} & \(\mathbf{R^2(Train)}\) \\
\hline
Spatter & 
$-47591.2675 - 0.1273 - 0.2616 P - 0.0237 V - 0.0944 P^2 + 0.2825 PV + 3.0756 V^2 + 0.2039 P^3 - 0.1063 P^2 V + 0.0093 PV^2 - 0.0102 V^3 - 0.0014 P^4 + 5.58 \times 10^{-4} P^3 V + 3.93 \times 10^{-5} P^2 V^2 - 1.76 \times 10^{-5} PV^3 + 1.47 \times 10^{-5} V^4 + 3.68 \times 10^{-6} P^5 - 1.14 \times 10^{-6} P^4 V - 2.27 \times 10^{-7} P^3 V^2 + 6.23 \times 10^{-8} P^2 V^3 - 2.26 \times 10^{-9} PV^4 - 8.52 \times 10^{-9} V^5 - 3.32 \times 10^{-9} P^6 + 1.26 \times 10^{-9} P^5 V - 2.78 \times 10^{-10} P^4 V^2 + 2.47 \times 10^{-10} P^3 V^3 - 8.39 \times 10^{-11} P^2 V^4 + 1.11 \times 10^{-11} PV^5 + 1.42 \times 10^{-12} V^6$ & 0.83 \\
\hline
Spatter & 
$-73673.5843 + 60.4532 L + 1719.4559 W + 424.9562 D + 1.6502 L^2 - 23.3099 LW + 14.7747 LD + 48.4069 W^2 - 77.2674 WD + 24.4086 D^2 - 0.0070 L^3 + 0.0752 L^2 W - 0.0317 L^2 D - 0.1816 LW^2 + 0.1530 LWD - 0.0400 LD^2 + 0.0515 W^3 + 0.0234 W^2 D - 0.0140 WD^2 - 0.0023 D^3$ & 0.8 \\
\hline
Spatter & 
$-1262141.6793 - 34.2697 \log L + 46.6761 W + 3.3128 D + 1832.5419 \log W - 19.5423 \log D + 2.0622 (\log L)^2 - 0.0189 \log L \cdot W + 6.79 \times 10^{-4} \log L \cdot D + 1.5434 \log L \cdot \log W + 1.1827 \log L \cdot \log D + 0.0032 W^2 + 0.0011 WD - 6.1771 W \cdot \log W - 0.1413 W \cdot \log D + 1.73 \times 10^{-5} D^2 - 0.3015 D \cdot \log W - 0.2274 D \cdot \log D - 347.0129 (\log W)^2 + 40.4521 \log W \cdot \log D - 26.0121 (\log D)^2$ & 0.85 \\
\hline
\end{tabular}
\end{table}

\FloatBarrier

\section{Conclusions}

In this work, we analyze a spatter dataset collected by coupling OpenFOAM and FLOW-3D, leveraging the strengths of both simulation packages. OpenFOAM generates realistic spatter occurrences but is computationally expensive, while FLOW-3D is computationally efficient but lacks the physics to produce realistic spatter phenomena. With this configuration, we collected a dataset comprising melt pool length, width, depth, cross-sectional area, volume, and volume indicated as spatter. The performance of machine learning models and polynomial regression was evaluated on the collected parameters, showing high accuracy of 95\% and above for melt pool dimensions and geometry features for both ML models and polynomial fitting. For predicting the volume indicated as spatter, the Extra-Trees model achieved the highest $R^2$ scores of 0.97 and 0.88 for the training and testing datasets, respectively. Using polynomial fitting, we derived equations for each melt pool dimension, geometry, and volume indicated as spatter based on model inputs, enhancing interpretability by showing how input variables influence the outputs. This findings open opportunities for robust control systems to monitor part production, minimize defect formation, and ensure repeatability in AM processes.

\section*{Declarations}
\subsection*{Funding}

This research was sponsored by the Army Research Laboratory and conducted under Cooperative Agreement Number W911NF-20-2-0175. The views and conclusions presented in this document are those of the authors and do not necessarily represent the official policies or positions of the Army Research Laboratory or the U.S. Government. The U.S. Government is authorized to reproduce and distribute reprints for governmental purposes, despite any copyright notice herein.

\subsection*{Competing Interests}

The authors have no competing interests to declare that are relevant to the content of this article.


\bibliography{reference}

\pagebreak
\appendix

\section{OpenFOAM simulation domain}
\begin{figure*}[htbp!]
\begin{center}
\includegraphics[width=1\linewidth]{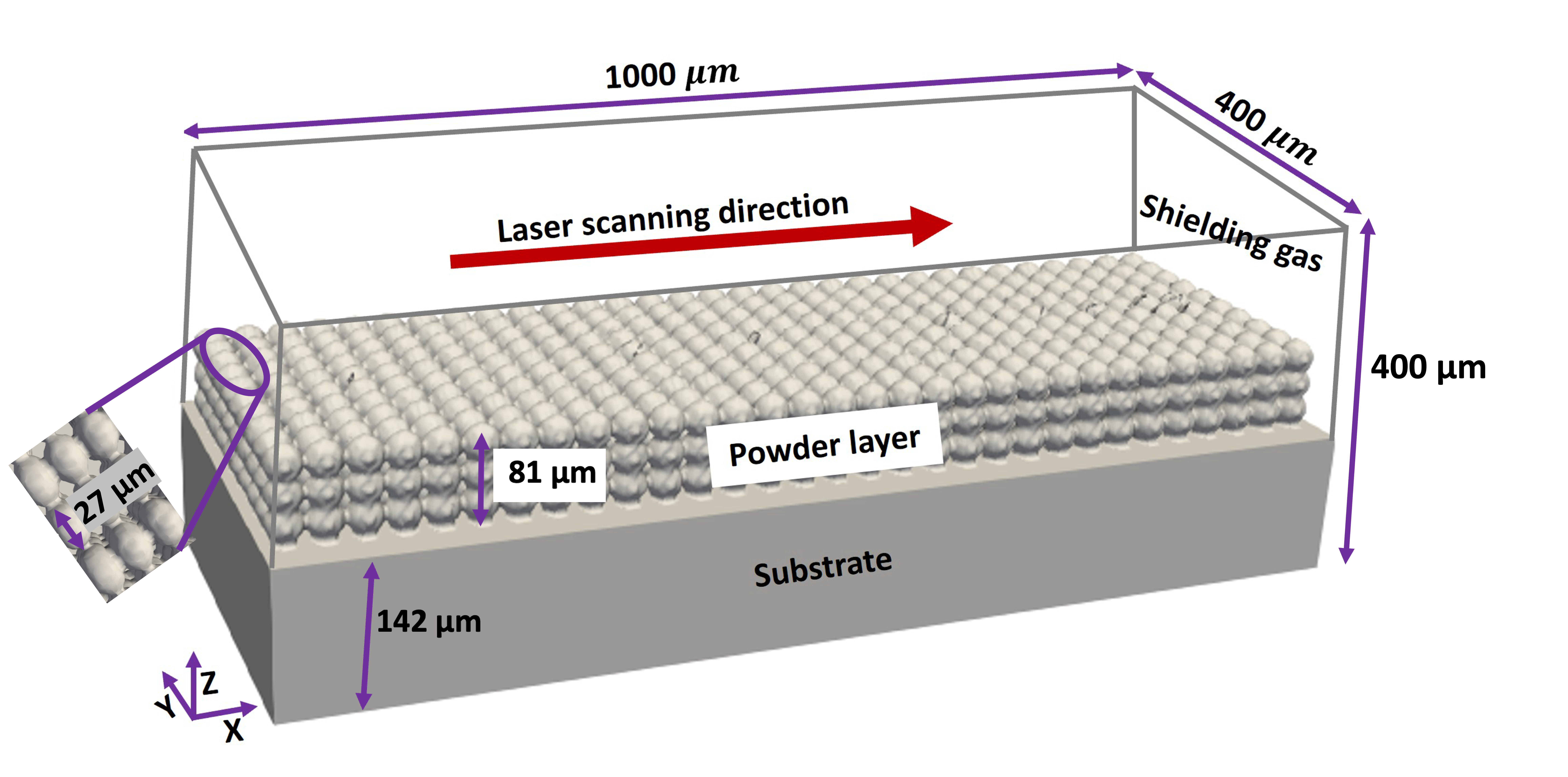}
\end{center}
    \caption{Schematic illustration of the computational domain with random packing of Stainless steel powder bed.}
\label{fig:fig1}
\end{figure*}

\subsection{Machine learning prediction}
\subsubsection{\normalfont Random Forest (RF)}
Random forest regression is an ensemble method that enhances predictive accuracy by averaging multiple decision trees, which reduces overfitting \cite{breiman2001random}. This technique leverages random sampling of data and features, capturing complex interactions and improving generalization \cite{hastie2009elements}. Known for its robustness in high-dimensional, nonlinear datasets, random forests also provide insights into feature importance, aiding model interpretability\cite{liaw2002classification}.
\subsubsection{\normalfont The Extra Trees Regressor}
The Extra Trees Regressor, or Extremely Randomized Trees, is an ensemble method similar to random forests but introduces more randomness to improve performance and reduce variance \cite{geurts2006extremely}. Instead of selecting optimal split points, it randomly chooses split thresholds for each feature, which leads to greater diversity among trees and often faster training. This approach enhances generalization and can handle complex, high-dimensional data efficiently, though it may require careful tuning to avoid underfitting in some cases \cite{pedregosa2011scikit}.
\subsubsection{\normalfont  Bagging Regressor}
The Bagging Regressor is an ensemble method that improves prediction accuracy by combining the outputs of multiple base regressors trained on randomly sampled subsets of the data \cite{breiman1996bagging}. This technique, also known as bootstrap aggregation, reduces model variance and enhances generalization by averaging the predictions of each model, often resulting in more robust performance compared to a single model \cite{hastie2009elements}. Bagging is particularly effective for high-variance models, such as decision trees, and can be customized with different base regressors to suit various tasks.
\subsubsection{\normalfont  The K-Nearest Neighbors (KNN) Regressor}
KNN Regressor is a non-parametric method that predicts the value of a target variable based on the average of its $k$ closest data points (neighbors) in the feature space \cite{altman1992introduction}. This approach is simple and effective for datasets where similar instances yield similar outputs, but its performance can be affected by high-dimensional data and may require scaling to handle features of different units effectively. The choice of $k$ impacts the model's bias-variance trade-off, with smaller values leading to more variance and larger values resulting in smoother, potentially biased predictions \cite{cover1967nearest}.
\subsubsection{\normalfont  The Gradient Boosting (GB) Regressor}
The Gradient Boosting Regressor is an ensemble learning method that builds a sequence of weak learners, typically decision trees, where each subsequent model corrects the errors of its predecessor \cite{friedman2001greedy}. By minimizing a loss function through gradient descent, it achieves high accuracy and handles complex, non-linear relationships well. Known for its robustness and predictive power, gradient boosting is effective in both small and large datasets, though it requires careful tuning to prevent overfitting and can be computationally intensive \cite{natekin2013gradient}.

\begin{table}[ht]
\centering
\caption{Polynomial regression equations for melt pool volume indicated as spatter, as the equation is derived from logarithm transformation of melt pool dimension }
\renewcommand{\arraystretch}{1.2}
\begin{tabular}{|c|>{\raggedright\arraybackslash}p{12cm}|c|}
\hline
\textbf{Features} & \textbf{Equation} & \(\mathbf{R^2(Train)}\) \\
\hline
Spatter & 
$-94877.9016 + 0.0016 P - 0.0113 V - 0.0131 \log V + 0.0192 P^2 - 0.0054 PV + 0.0023 P \log V - 0.0043 V^2 - 2.81 \times 10^{-4} V \log V - 6.35 \times 10^{-4} (\log V)^2 + 0.3343 P^3 + 0.0824 P^2 V + 0.0351 P^2 \log V - 0.0297 PV^2 - 0.0093 PV \log V + 1.98 \times 10^{-4} P (\log V)^2 - 0.0378 V^3 - 0.0140 V^2 \log V - 0.0019 V (\log V)^2 - 1.05 \times 10^{-4} (\log V)^3 - 0.0303 P^4 + 0.7149 P^3 V + 0.2123 P^3 \log V + 0.0801 P^2 V^2 + 0.0937 P^2 V \log V + 0.0668 P^2 (\log V)^2 - 0.1631 PV^3 - 0.0153 PV^2 \log V - 0.0357 PV (\log V)^2 - 0.0020 P (\log V)^3 + 0.0014 V^4 - 0.0473 V^3 \log V - 0.0347 V^2 (\log V)^2 - 0.0075 V (\log V)^3 - 5.44 \times 10^{-4} (\log V)^4 + 6.57 \times 10^{-6} P^5 + 2.25 \times 10^{-4} P^4 V + 0.0143 P^4 \log V - 7.38 \times 10^{-4} P^3 V^2 - 0.2463 P^3 V \log V - 0.6532 P^3 (\log V)^2 + 1.34 \times 10^{-4} P^2 V^3 - 0.0214 P^2 V^2 \log V + 0.0456 P^2 V (\log V)^2 + 0.0618 P^2 (\log V)^3 - 1.87 \times 10^{-5} PV^4 + 0.0427 PV^3 \log V + 0.1350 PV^2 (\log V)^2 - 0.0936 PV (\log V)^3 - 0.0160 P (\log V)^4 - 1.04 \times 10^{-6} V^5 - 0.0018 V^4 \log V - 0.0172 V^3 (\log V)^2 - 0.0543 V^2 (\log V)^3 - 0.0257 V (\log V)^4 - 0.0026 (\log V)^5 + 2.08 \times 10^{-9} P^6 + 6.55 \times 10^{-9} P^5 V - 2.34 \times 10^{-6} P^5 \log V + 8.67 \times 10^{-9} P^4 V^2 - 2.98 \times 10^{-5} P^4 V \log V - 0.0020 P^4 (\log V)^2 - 1.53 \times 10^{-8} P^3 V^3 + 8.50 \times 10^{-5} P^3 V^2 \log V + 0.0244 P^3 V (\log V)^2 + 0.0579 P^3 (\log V)^3 + 5.18 \times 10^{-9} P^2 V^4 - 1.74 \times 10^{-5} P^2 V^3 \log V + 0.0010 P^2 V^2 (\log V)^2 - 0.0073 P^2 V (\log V)^3 - 0.1802 P^2 (\log V)^4 - 1.17 \times 10^{-9} PV^5 + 2.95 \times 10^{-6} PV^4 \log V - 0.0029 PV^3 (\log V)^2 + 0.0018 PV^2 (\log V)^3 - 0.0679 PV (\log V)^4 - 0.0785 P (\log V)^5 + 9.03 \times 10^{-11} V^6 + 1.59 \times 10^{-8} V^5 \log V + 2.22 \times 10^{-4} V^4 (\log V)^2 + 0.0071 V^3 (\log V)^3 + 0.0482 V^2 (\log V)^4 - 0.0750 V (\log V)^5 - 0.0110 (\log V)^6$ & 0.9 \\
\hline
\end{tabular}
\end{table}
\FloatBarrier

\end{document}